\documentclass[a4paper, 10pt, conference]{cssconf}      

\IEEEoverridecommandlockouts                              

\usepackage{graphics} 
\usepackage{graphicx}
\graphicspath{{fig/}}
\usepackage{multirow}

\title{\LARGE \bf
Baseline CNN structure analysis for facial expression recognition
}

\author{Minchul Shin$^{1}$, Munsang Kim$^{2}$ and Dong-Soo Kwon$^{1}$
\thanks{*This work was supported by the Industrial Strategic Technology Development Program (10044009, Development of a self-improving bidirectional sustainable HRI technology) funded by the Ministry of Knowledge Economy (MKE, Korea)}
\thanks{$^{1}$Minchul Shin is with the Department of Mechanical Engineering and Human-Robot Interaction Research Center, Korea Advanced Institute of Science and Technology (KAIST), Daejeon 305-701, Republic of Korea (phone: +82-42-350-8212; fax: +82-42-350-8240).
        {\tt\small min.stellastra@gmail.com}}%
\thanks{$^{1}$Dong-Soo Kwon is with the Department of Mechanical Engineering and Human-Robot Interaction Research Center, KAIST, Daejeon 305-701, Republic of Korea (phone: +82-42-350-3042; fax: +82-42-350-8240).
        {\tt\small kwonds@kaist.ac.kr}}%
}

\begin{document}

\maketitle
\thispagestyle{empty}
\pagestyle{empty}

\begin{abstract}
We present a baseline convolutional neural network (CNN) structure and image preprocessing methodology to improve facial expression recognition algorithm using CNN. To analyze the most efficient network structure, we investigated four network structures that are known to show good performance in facial expression recognition. Moreover, we also investigated the effect of input image preprocessing methods. Five types of data input (raw, histogram equalization, isotropic smoothing, diffusion-based normalization, difference of Gaussian) were tested, and the accuracy was compared. We trained 20 different CNN models (4 networks x 5 data input types) and verified the performance of each network with test images from five different databases. The experiment result showed that a three-layer structure consisting of a simple convolutional and a max pooling layer with histogram equalization image input was the most efficient. We describe the detailed training procedure and analyze the result of the test accuracy based on considerable observation.
\end{abstract}


\section{INTRODUCTION}

Human facial expression recognition is considered important in the human-robot interaction field and has been studied with much interest in the past 10 years \cite{c1,c2,c3}. However, despite the enormous efforts, it still remains as a challenging task for robots. The traditional approach for facial expression recognition consists of two main parts: feature extraction and classification. Features extracted from the training data always play an important role in the recognition problem because the classifier makes its decision based on the combination of extracted features. Handcrafted features such as LBP, HOG, and SIFT have been widely used in the traditional approach owing to their proven performances under specific circumstances and to their low computational cost for feature extraction process \cite{c4,c5,c6}. Shan {\it et al.} \cite{c4} formulated a boosted-LBP feature and combined it with a support vector machine (SVM) classifier. Their method performed robustly and stably over a useful range of low-resolution facial images. Berretti {\it et al.} \cite{c5} computed the SIFT descriptor on 3D facial landmarks of depth images and used SVM for the classification. Albiol {\it et al.} \cite{c6} proposed an HOG descriptor-based EBGM algorithm. Gabor features from EBGM were replaced by HOG descriptors, achieving good performance. Although these approaches reported good accuracy, the handcrafted feature has its inherent drawbacks. When we use handcrafted features, either unintended features that have no effects on classification may get included or important features that have a great influence on the classification may get omitted. This is because the features are ``crafted'' by human experts, and the experts may not be able to consider all possible cases and include them in the feature.

With recent advances in deep learning and parallel computing, applying convolutional neural network (CNN)-based deep neural networks into a classification problem has attained impressive successes \cite{c7,c8,c9,c10,c11,c12,c13}. Deep learning methods are distinguishable from traditional machine learning algorithms in that they perform the feature extraction and classification process simultaneously. Another advantage of using deep learning methods is that, since they extract features through iterative weight update by backpropagation and error optimization, the classifier could include critical and unforeseen features that humans hardly come up with. This process is called feature learning, and CNN is especially suitable for processing 2D image-based training datasets. CNN can be seen as a special type of multilayer perceptron (MLP), but CNN rather focuses on the local relationships between pixels by using receptive fields. According to Goodfellow {\it et al.} [7], the performance of a feature learning algorithm can be proven through the result of a facial expression recognition competition (FER-2013). The top three teams out of 56 teams that participated in the aforementioned competition all used CNNs, and the result showed that the features learned by the CNN are indeed capable of outperforming handcrafted features, although the difference is not extreme.

In this paper, we present a comparison between various types of CNN structures, and find out the most effective structure for  application in facial expression recognition. Four types of CNN structure were tested, and the one that showed the highest accuracy was selected as a target structure for fine parameter tuning. In addition to the CNN structure selection, we also tested five types (raw, histogram-equalized, isotropic diffusion-based normalization, and LBP) of differently preprocessed images to verify the most suitable form of training images. We expect that, by setting up the baseline CNN structure and image preprocess method, the recognition accuracy of other CNN-based deep learning approach will improve.

\section{RELATED WORKS}
Facial expression recognition with CNN-based deep learning was reported in Refs. \cite{c8,c9,c10,c11,c12,c13}. Tang, the winner of the ICMLW2013 facial expression recognition challenge, reported that the implementation of a multi-class L2-SVM instead of a softmax layer for loss calculation was actually able to improve the classification accuracy \cite{c8}. Since L2-SVM is in quadratic form, which is differentiable, switching softmax to SVM is simple and shows a slight increase in accuracy compared to softmax. Liu {\it et al.} \cite{c9} used 3D-CNN and a deformable facial action part model to localize facial action parts and learn part-based features for video-based emotion classification. The kernel size can also be an important factor for CNN performance. Fasel \cite{c10} implemented an intentionally large receptive field (11x11) for integrating the features found by the early stage convolutional layer. The classifier ensemble technique is frequently used for boosting the classification performance \cite{c11,c12,c13}. Yu {\it et al.} \cite{c11} reported that the random initialization of neural networks not only leads to varying network parameters, but also renders the classification ability of diverse networks. For this reason, the ensemble technique usually shows concrete performance improvement. Kahou {\it et al.} \cite{c12} combined modality-specific multiple deep neural networks. Their method mixed a number of modalities such as facial image, audio, bag of mouth features with CNN, and deep restricted Boltzmann machine, and the final predictions of each classifier were averaged. Yu {\it et al.} \cite{c11} independently trained multiple differently initialized CNNs and tested the log-likelihood loss and hinge loss for the ensemble optimization method of output responses. Their method resulted in a slight increase in accuracy compared to the single CNN model. Lastly, Kim {\it et al.} \cite{c13} proposed a hierarchical committee of deep CNNs with exponentially weighted decision fusion. They trained 216 CNN models sharing the same structure, but with different weights and ensembled them with their own decision fusion rule called VA-Expo-WA. This approach requires a large amount of computing power, but showed a fine performance, winning the EmotiW2015 facial expression recognition competition. As it can be seen so far, plenty of papers for facial expression recognition used CNN-based deep neural networks, and their baseline structures and image preprocessing methods were all different. The reason is that, in most cases, the CNN structure selection or data preprocessing method is not their main research focuses. However, choosing the proper CNN structure and preprocessed input image type is important for improving accuracy. In this paper, we will compare five preprocessed image types and four types of CNN structure and suggest the most suitable baseline CNN model and image type for facial expression recognition.

\section{DATA PREPARATION}
In this section, we first describe what type of dataset is used for the experiment and how it is modified. Then, the face detection method and face registration processes will be discussed in sequence.
\subsection{Dataset}
To train a deep convolutional network and verify its performance, we used multiple kinds of datasets. For the training dataset, the Facial Expression Recognition 2013 (FER-2013) dataset released for the ICMLW subchallenge and the Static Facial Expression in the Wild (SFEW2.0) dataset released for the EmotiW2015 competition were used. FER-2013 contains 28,709 training faces, 3589 private test faces, and 3589 public test faces. We used training faces and public test faces for the training and left private test faces for the accuracy test. FER-2013 facial images are not exactly frontal, including a variation of the rotation and transition with a 48x48 grayscale format. Since FER-2013 images were collected using the Google Image Search API, they contain a variety of facial expressions existing in real-world conditions. The SFEW2.0 dataset consists of 944 training faces, 422 validation faces, and 372 test faces. The SFEW dataset is a static subset of AFEW, which contains video clips extracted from movies. Although the emotions in movies are not very spontaneous, they provide facial expressions in a much more natural and versatile way than those found in laboratory-controlled datasets \cite{c11}.

For the performance evaluation, five different datasets were chosen: FER-2013, SFEW2.0, CK+ (extended Cohn-Kanade), KDEF (Karolinska Directed Emotional Faces), and Jaffe \cite{c14, c15, c16}. CK+ is the most widely used dataset for facial expression recognition, and it includes seven different facial expression labels. KDEF contains facial images of 70 individuals and 7 different facial expressions from 5 different angles. This dataset was originally developed for use in psychological and medical research purposes, but is also suitable for facial expression recognition. Jaffe contains facial images of Japanese females, and the head pose is almost frontal. Since there are very few datasets containing facial images of Asians, Jaffe can be a good test set for performance evaluation. All datasets used in this work included seven types of the same emotion expression, and every single image was fully labeled. 

   \begin{figure}[t]	
   	\flushleft
   	\includegraphics[scale=0.23]{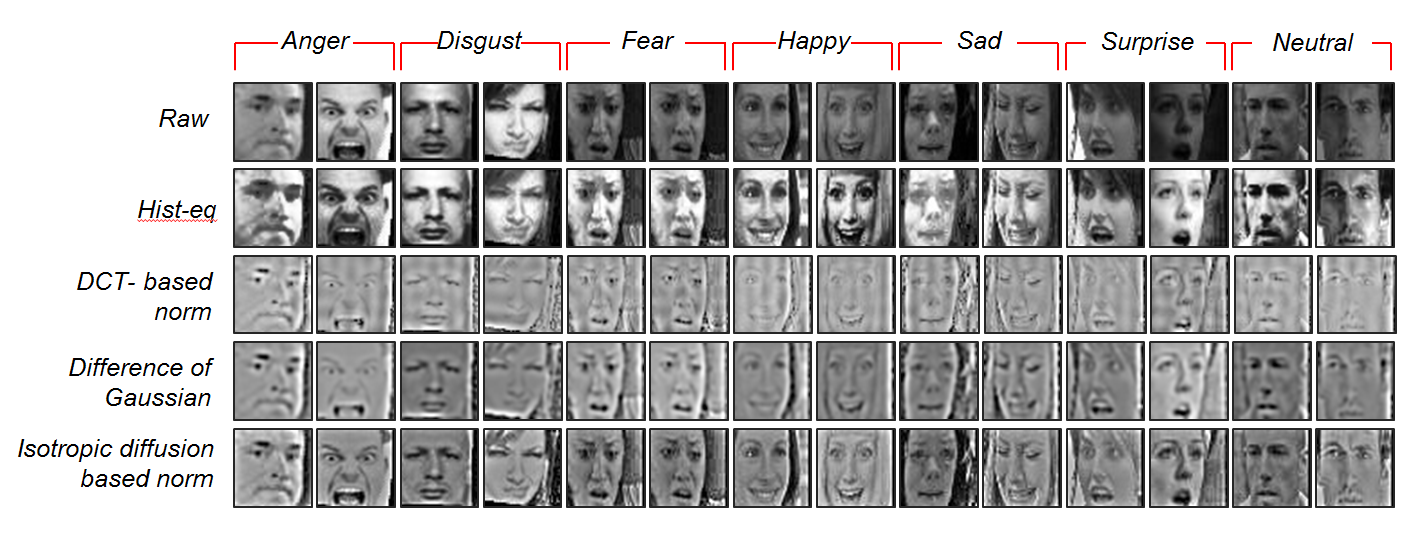}
   	\caption{Five different image preprocessing methods applied for the aligned training faces.}
   	\label{figurelabel}
   	
   	\flushleft
   	\includegraphics[scale=0.3]{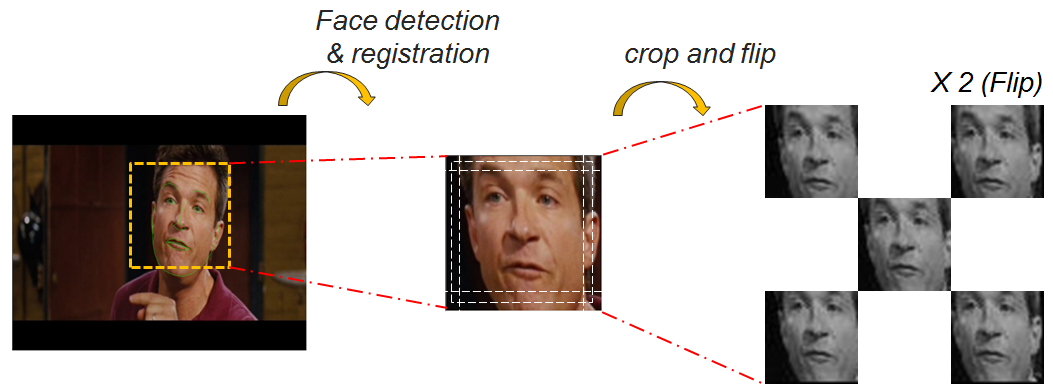}
   	\caption{The entire procedure for face detection, registration, cropping, and flipping (center, top right, top left, bottom right, and bottom left).}
   	\label{figurelabel}
   \end{figure}

\subsection{Face Detection and Registration}
Most of the papers on the subject used the face crop technique owing to the robustness of its performance and to its high accuracy. We also cropped faces from each dataset and aligned the faces with respect to the landmark position of the eye. Every single facial image was rotated so that the straight line connecting two eye positions becomes parallel to the horizontal line. For this task, we needed a face detector that could extract the face landmark, and the dlib C++ library provided by King \cite{c20} performed very well for this purpose. The face detection algorithm implemented in dlib is based on Ref. \cite{c21} with fast response. Kazemi and Sullivan \cite{c21} proposed a face alignment method based on the ensemble of regression trees that performs shape invariant feature selection. A simple example code is provided so that anyone can easily try and test the method. Faces were detected and cropped from the facial images in the datasets, and the number of images that were successfully processed is shown in Table 1. Det\# stands for the number of successful detections, Manual\# for the number of images left after manual filtering by the authors, and Crop\# for the number of images after cropping and flipping. The success rate of the face detection differed depending on the dataset, with the highest rate, 0.802, for SFEW2.0 and the lowest rate, 0.607, for KDEF. There are two reasons for the failure of the face detection. First, the face detector could not find faces if only one side of the face is shown in the picture. This is because the ensemble model of the face detector in dlib was trained with the Helen dataset, in which training images are close to frontal face. Since the images in the KDEF dataset were captured at five different angles, plenty of faces were not frontal. Second, the FER-2013 dataset provides cropped facial images from the beginning, where the faces are fully filled in 48x48 pixels. For this reason, the face landmark detector often failed to extract the landmark, and failed images were excluded from the training data to eliminate the possible effects from face rotation. Although 68.53\% of the faces in total were detected and processed, more than 25K training images were enough to include various features of facial expression, and the result will be discussed later. 
\begin{table}[h]
	\caption{Number of correct detections for each dataset}
	\label{table_example}
	\begin{center}
		\begin{tabular}{|c|c|c|c|c|c|}
			\hline
			& FER-2103 & SFEW2.0 & CK+ & KDEF & Jaffe\\
			\hline
			Face \# & 35,887 & 1,366 & 309 & 4,898 & 181 \\
			\hline
			Det \#  & 24,668 & 1,095 & 309 & 2972 & 181\\
			\hline
			Rate (Det\#/Face\#) & 0.687 & 0.802 & 1.000 & 0.607 & 1.000 \\
			\hline
			Manual \# & 24,657 & 832 & 308 & 2961 & 180\\
			\hline
			Crop  \#  & 246,570 & 8320 & 3080 & 29610 & 1800  \\
			\hline
		\end{tabular}
	\end{center}
\end{table}

   \begin{figure}[h]	
   	\flushleft
   	\includegraphics[scale=0.27]{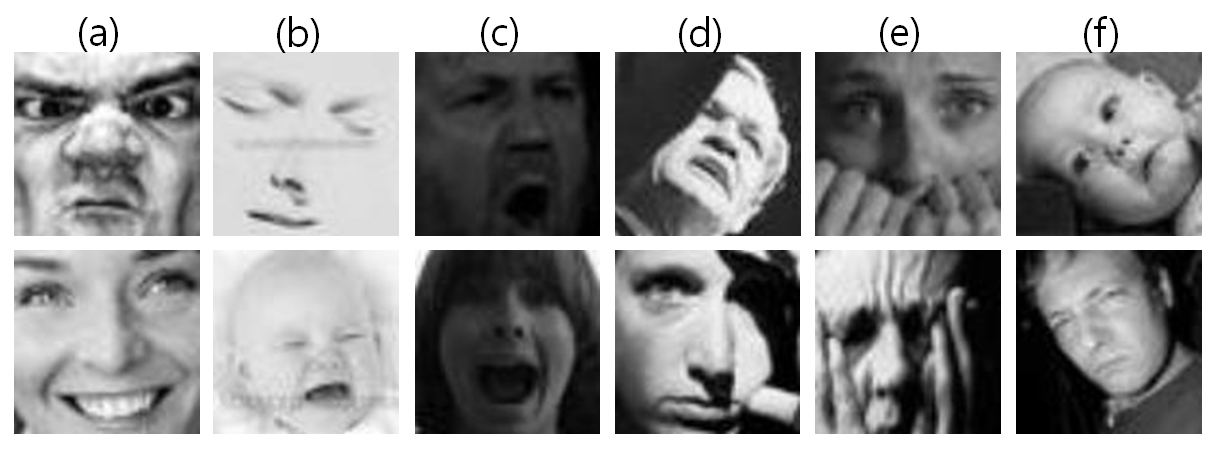}
   	\caption{The failed cases of face detection; (a)excessively filled (b)unclear edge (c)dark contrast (d)occulusion by shadow (e)partially visible (f)rotated.}
   	\label{figurelabel}
   \end{figure}

\begin{table*}[t]
	\caption{Accuracy result}
	\begin{center}
		\begin{tabular}{|p{0.8cm}|p{1.2cm}|p{1cm}|p{0.7cm}|p{1cm}|p{0.7cm}|p{1cm}|p{0.7cm}|p{1cm}|p{0.7cm}|}
			\hline
			\multicolumn{2}{|c|}{\multirow{2}{*}{Test set (\%)}} & \multicolumn{2}{|c|}{Tang} & \multicolumn{2}{|c|}{Yu} & \multicolumn{2}{|c|}{Kahou} & \multicolumn{2}{|c|}{Caffe-ImageNet} \\ \cline{3-10}
			\multicolumn{2}{|c|}{ } & Indiv. & Avg. &  Indiv. & Avg. &  Indiv. & Avg. &  Indiv. & Avg.\\
			\hline
			\multirow{5}{*}{Raw} & FER-2013 & 62.20 & \multirow{5}{*}{(35)} & 61.87 & \multirow{5}{*}{(35)} & 58.35 & \multirow{5}{*}{(70)} & 60.58 & \multirow{5}{*}{(30)} \\
			\cline{2-3} \cline{5-5} \cline{7-7} \cline{9-9}
			& SFEW2.0 & 45.96 & 54.58 & 50.6 & 54.70 & 45.25 & 52.58 & 51.22 & 55.09 \\
			\cline{2-3} \cline{5-5} \cline{7-7} \cline{9-9}
			& CK+ & 60.98 &  & 59.93 &  & 56.72 &  & 59.02 & \\
			\cline{2-3} \cline{5-5} \cline{7-7} \cline{9-9}
			& KDEF & 54.18 &  & 53.21 &  & 52.28 &  & 56.86 & \\
			\cline{2-3} \cline{5-5} \cline{7-7} \cline{9-9}
			& Jaffe & 49.56 &  & 47.89 &  & 50.28 &  & 47.78 & \\
			\hline
			
			\hline
			\multirow{5}{*}{Hist-eq} & FER-2013 & 66.67 & \multirow{5}{*}{(30)} & 64.98 & \multirow{5}{*}{(35)} & 65.97 & \multirow{5}{*}{(70)} & 66.99 & \multirow{5}{*}{(30)} \\
			\cline{2-3} \cline{5-5} \cline{7-7} \cline{9-9}
			& SFEW2.0 & 64.84 & 59.38  & 66.65 & 57.39  & 68.14 & 58.31 & 68.46 & 59.29 \\
			\cline{2-3} \cline{5-5} \cline{7-7} \cline{9-9}
			& CK+ & 65.54 &  & 60.92 &  & 57.48 &  & 63.11 & \\
			\cline{2-3} \cline{5-5} \cline{7-7} \cline{9-9}
			& KDEF & 50.66 &  & 49.5 &  & 50.02 &  & 51.12 & \\
			\cline{2-3} \cline{5-5} \cline{7-7} \cline{9-9}
			& Jaffe & 49.17 &  & 44.89 &  & 49.94 &  & 46.78 & \\
			\hline
			
			\hline
			\multirow{5}{*}{IS} & FER-2013 & 62.16 & \multirow{5}{*}{(35)} & 62.82 & \multirow{5}{*}{(32)} & 61.34 & \multirow{5}{*}{(70)} & 62.73 & \multirow{5}{*}{(23)} \\
			\cline{2-3} \cline{5-5} \cline{7-7} \cline{9-9}
			& SFEW2.0 & 52.9 & 56.16  & 54.35 & 57.65  & 58.09 & 56.93 & 53.90 & 58.58 \\
			\cline{2-3} \cline{5-5} \cline{7-7} \cline{9-9}
			& CK+ & 62.26 &  & 63.8 &  & 61.77 &  & 66.49 & \\
			\cline{2-3} \cline{5-5} \cline{7-7} \cline{9-9}
			& KDEF & 57.28 &  & 58.93 &  & 56.12 &  & 59.15 & \\
			\cline{2-3} \cline{5-5} \cline{7-7} \cline{9-9}
			& Jaffe & 46.22 &  & 48.33 &  & 47.33 &  & 50.61 & \\
			\hline
			
			\hline
			\multirow{5}{*}{DCT} & FER-2013 & 56.09 & \multirow{5}{*}{(40)} & 58.33 & \multirow{5}{*}{(50)} & 54.08 & \multirow{5}{*}{(70)} & 56.15 & \multirow{5}{*}{(22)} \\
			\cline{2-3} \cline{5-5} \cline{7-7} \cline{9-9}
			& SFEW2.0 & 46.69 & 51.50 & 51.02 & 53.52 & 52.88 & 50.77 & 45.92 & 50.77 \\
			\cline{2-3} \cline{5-5} \cline{7-7} \cline{9-9}
			& CK+ & 54.33 &  & 57.48 &  & 53.60 &  & 53.54 & \\
			\cline{2-3} \cline{5-5} \cline{7-7} \cline{9-9}
			& KDEF & 54.92 &  & 52.96 &  & 48.88 &  & 54.30 & \\
			\cline{2-3} \cline{5-5} \cline{7-7} \cline{9-9}
			& Jaffe & 45.5 &  & 47.83 &  & 44.39 &  & 43.94 & \\
			\hline
						
			\hline
			\multirow{5}{*}{DOG} & FER-2013 & 58.96 & \multirow{5}{*}{(35)} & 59.38 & \multirow{5}{*}{(40)} & 57.84 & \multirow{5}{*}{(60)} & 58.94 & \multirow{5}{*}{(25)} \\
			\cline{2-3} \cline{5-5} \cline{7-7} \cline{9-9}
			& SFEW2.0 & 49.37 & 53.35 & 50.47 & 52.52 & 57.82 & 53.28 & 52.17 & 54.05 \\
			\cline{2-3} \cline{5-5} \cline{7-7} \cline{9-9}
			& CK+ & 56.03 &  & 55.28 &  & 56.75 &  & 57.25 & \\
			\cline{2-3} \cline{5-5} \cline{7-7} \cline{9-9}
			& KDEF & 55.19 &  & 56.3 &  & 48.40 &  & 54.38 & \\
			\cline{2-3} \cline{5-5} \cline{7-7} \cline{9-9}
			& Jaffe & 47.05 &  & 41.17 &  & 45.61 &  & 47.50 & \\
			\hline 
			
			\multicolumn{10}{l}{\footnotesize \textit{a} Average indicates average accuracy (selected model epoch through the validation)}
			
		\end{tabular}
	\end{center}
\end{table*}

\begin{table*}[t]
	\caption{Network structure descriptions}
	\begin{center}
		\begin{tabular}{|p{0.8cm}|p{0.74cm}|p{0.78cm}|p{0.78cm}|p{0.78cm}|p{0.78cm}|p{0.75cm}|p{0.78cm}|p{0.78cm}|p{0.8cm}|p{0.73cm}|p{0.78cm}|p{0.73cm}|p{0.70cm}|p{0.65cm}|}
			\hline
			network&category&layer1&layer2&layer3&layer4&layer5&layer6&layer7&layer8&layer9&layer10&layer11&layer12&layer13 \\
			\hline
			\multirow{3}{*}{Tang}&layer&conv&maxp&conv&maxp&conv&maxp&fc&output&&&&& \\
			\cline{2-15}
			&kernel&5x5(1,2)&3x3(2,1)&4x4(1,1)&3x3(2,1)&5x5(1,2)&3x3(2,1*)&-&-&&&&& \\
			\cline{2-15}
			&maps&42@32&21@32&20@32&10@32&42@32&42@32&1@3072&1@7&&&&& \\
			\hline
			\multirow{3}{*}{Yu}&layer&conv&stochp&conv&conv&stochp&conv&conv&stochp&fc&fc&output&& \\
			\cline{2-15}
			&kernel&5x5(1,2)&3x3(2,1)&3x3(1,1)&3x3(1,1)&3x3(2,1)&3x3(1,1)&3x3(1,1)&3x3(2,0)&-&-&-&& \\
			\cline{2-15}
			&maps&42@48&21@48&21@48&11@64&11@128&11@128&5@128&1@1024&1@1024&1@7&&& \\
			\hline
			\multirow{3}{*}{Kahou}&layer&conv&maxp&lrn&conv&avgp&lrn&conv&avgp&fc&output&&& \\
			\cline{2-15}
			&kernel&5x5(1,2)&3x3(2,0)&-&3x3(1,1)&3x3(2,1)&-&3x3(1,1)&3x3(2,1*)&-&-&&& \\
			\cline{2-15}
			&maps&42@64&21@64&21@64&20@64&10@64&10@128&5@128&1@3072&1@7&&&& \\
			\hline
			\multirow{3}{*}{ImageNet}&layer&conv&maxp&lrn&conv&maxp&lrn&conv&conv&conv&maxp&conv&fc&output \\
			\cline{2-15}
			&kernel&5x5(1,2)&3x3(2,0)&-&3x3(1,1)&3x3(2,1)&-&3x3(1,1)&3x3(1,1)&3x3(1,1)&3x3(2,1*)&5x5(1,0)&-&- \\
			\cline{2-15}
			&maps&42@32&20@32&20@32&20@96&10@96&10@96&10@128&10@128&10@96&5@96&1@1024&1@1024&1@7 \\
			\hline
			
			\multicolumn{15}{l}{\footnotesize \textit{a} conv, fc, lrn, maxp, avgp, stochp: convolutional, fully-connected, local response norm, max-pooling, average-pooling, stochastic-pooling} \\
			\multicolumn{15}{l}{\footnotesize \textit{b} kernel: [kernel size]([stride],[zero-padding]) where 'padding with asterik*' refers to zero-padding to top, left direction only. } \\
			\multicolumn{15}{l}{\footnotesize \textit{c} maps: [size of output maps]@[the nubmer of output maps]}
			
		\end{tabular}

	\end{center}
\end{table*}

\subsection{Generating transition variation}
Yu {\it et al.} \cite{c11} reported that the random perturbation through cropping of images essentially generates additional unseen training samples and therefore makes the network more robust. A similar image cropping technique was also applied in Refs. \cite{c12,c13}. The images were cropped and flipped for transition variation in both papers. On the basis of this observation, we also cropped the original 48x48 facial images into a size of 42x42 of five crops (center, top left, top right, bottom left, and bottom right) and flipped them. Consequently, the training data were augmented by 10 times. In total, 227,890 face images were reproduced and used for the training set. 

\section{IMAGE PREPROCESSING METHOD}
Illumination or contrast influences the result accuracy greatly, depending on how it is dealt with. In this paper, five different frequently used face preprocessing methods (raw, histogram equalization, isotropic diffusion-based normalization, DCT-based normalization, difference of Gaussian) were tested. Histogram equalization (Hist-eq) is a contrast enhancement technique that usually increases the global contrast of images. This method is effective when the background's and foreground's brightness are almost the same. Isotropic diffusion-based normalization, so-called isotropic smoothing (IS), is a technique that aims to reduce image noise without removing significant parts of the image content such as edges or lines. The discrete cosine transform (DCT)-based normalization technique was proposed by Maheshkar {\it et al.} \cite{c17} and has been popularly applied to facial images owing to its powerful transform ability in image processing. Lastly, difference of Gaussian (DoG) is the most basic feature enhancement technique, which involves the subtraction of two differently blurred images. Usually, DoG is very powerful for increasing the visibility of edges and for representing texture details. Hist-eq was applied using the OpenCV library, and the IS, DCT, and DoG were applied using the transform function provided by the INface toolbox with default parameter settings. 

\section{CNN STRUCTURE CANDIDATES}
For the candidate of baseline CNN structure for facial expression, we chose four different structures based on other researchers' works. The first one is Tang's CNN structure \cite{c8}, which was also used as a baseline structure in KIM's work. It consists of one input transform and three convolutional and pooling layers, followed by a fully connected two-layer MLP. The next is Yu's structure \cite{c11}, which contains five convolutional layers, three stochastic pooling layers, and three fully connected layers. The network has two convolutional layers prior to pooling, except for the first layer. The third one is Kahou's structure \cite{c12}. It consists of three convolutional pooling layers and a two-layer MLP, with local response normalization (spatial batch normalization) and average pooling. The last candidate is the Caffe-ImageNet structure \cite{c19}. This structure was designed to classify 1000 classes from the ImageNet dataset, but in this paper, the number of last output nodes was reduced to seven. For all  four candidates, a rectified linear unit (ReLU) layer and a dropout layer were applied to every single convolutional and fully connected layers. It is generally believed that dropout prevents the network from being overfitted, and ReLU enables deep neural networks to be trained readily without additional initialization or a normalization process such as an autoencoder or a restricted Boltzmann machine \cite{c18}. The detailed parameter descriptions of the structure are described in Table 3, where the kernel category is represented as \{filter size\}(\{stride\},\{padding\}).

\section{EXPERIMENT}
\subsection{Network classification performance}
The classification performance result of each network is shown in Table 2. Tests were performed on five test sets (FER-2013, SFEW2.0, CK+, KDEF, and Jaffe) with four different network structures (Tang, Yu, Kahou, and Caffe-ImageNet). As can be seen in Figure 4, the Hist-eq method showed the highest performance for all four different network candidates. According to Table 2, the average accuracy of Hist-eq is the highest followed by IS, except for Yu's structure. It is interesting to note that the raw method generally showed better validation accuracy than IS did, as shown in Fig. 3, but the test accuracy showed better performance for IS. It indicates that the network is sensitive to illumination or contrast variation, such that reducing the illumination or contrast influences becomes more important for classifying test images that were not included in the training set. In addition, we can see that Hist-eq outperformed the other preprocessing methods, specifically for the SFEW test set. The reason is not clear, but we guess that contrast enhancement techniques such as Hist-eq fit because the SFEW database contains extracted images from movie scenes, which usually include a big contrast. From this observation, we concluded that the Hist-eq method is the most reliable for facial expression application.
Suppose that Hist-eq was chosen, Tang's model showed the best performance and resulted in a 59.38\% test accuracy. Although Caffe-Imagenet achieved an almost similar accuracy to Tang's model, the network complexity is different. Caffe-Imagenet was originally designed for classifying 1000 objects, so it has much a more complex structure and nodes. In practice, Caffe-Imagenet consumed almost twice as much computing power to complete the training. Despite the complex network structure, the fact that the test accuracy of Tang's model and Caffe-Imagenet was similar means that both networks were complex enough to include the features extractable from a 42x42 input dimension. In this respect, choosing Tang's network seems reasonable, saving more computing power. The best baseline CNN structure can vary depending on the image preprocessing methods. For example, if we use a DCT-based normalization technique, the best structure could be Yu's structure. We suggest carefully choosing the network structure and the image preprocessing method by referring to the result shown in Table 2. 

   \begin{figure}[h]	
   	\flushleft
   	\includegraphics[scale=0.52]{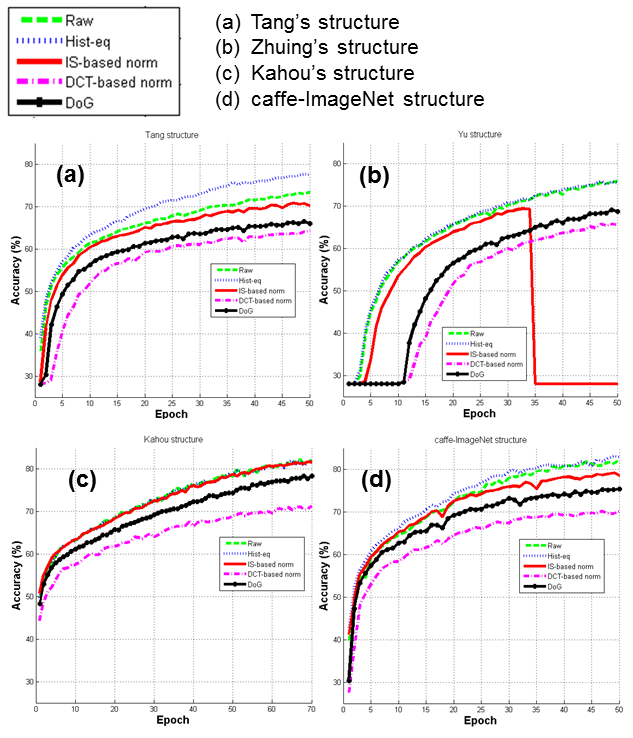}
   	\caption{Validation accuracy of the five different preprocessing methods.}
   	\label{figurelabel}
   \end{figure}

\subsection{Train and validation details}
We used the Torch7 deep learning library on an NVIDIA GeForce GTX 750 Ti GPU. The loss calculation for weight backpropagation was performed based on a stochastic gradient descent method, with a batch size of 50 and a momentum of 0.9. A fixed learning rate (0.005) was applied and was not changed during the iteration. No fine-tuning was done previously, and weight decay was fixed to 0.00001. To avoid overfitting, we added a dropout layer to every single convolutional and fully connected layer, and to the ReLU layer as well. The initial weight was set with the Xavier method, and a softmax layer was used for the output layer.
When training, the validation was performed in every epoch to avoid overfitting. In every epoch, 10\% of the training images were randomly chosen for validation, and the other images were used for training.

\section{CONCLUSIONS}
In this paper, we investigated a more efficient network structure and data preprocessing method for establishing a baseline structure for facial expression recognition. For the preprocessing method of the input image, the Hist-eq method showed the most reliable performance for all the network models. Moreover, we found that Tang's network achieved reasonably high accuracy with Hist-eq images compared to the other networks, even with less network complexity. On the basis of this observation, we suggest Tang's simple network with Hist-eq images as a baseline CNN model for further research. We expect that this baseline structure can help any trials of CNN-based algorithms that use the ensemble technique such as committee machines or any other research using a single CNN structure to choose a reasonable network structure.

\addtolength{\textheight}{-11cm}

\section*{ACKNOWLEDGMENT}
This work was supported by the Industrial Strategic Technology Development Program (10044009, Development of a self-improving bidirectional sustainable HRI technology) funded by the Ministry of Knowledge Economy (MKE, Korea)

\end{document}